\pdfoutput=1
\documentclass{article}

\usepackage{PRIMEarxiv}

\usepackage[utf8]{inputenc} 
\usepackage[T1]{fontenc}    
\usepackage{hyperref}       
\usepackage{url}            
\usepackage{booktabs}       
\usepackage{amsfonts}       
\usepackage{nicefrac}       
\usepackage{microtype}      
\usepackage{lipsum}
\usepackage{fancyhdr}       
\usepackage{graphicx}       
\graphicspath{{media/}}     
\usepackage{float}
\usepackage{amsmath}
\usepackage{amssymb}
\usepackage[]{multirow}
\usepackage[table,xcdraw]{xcolor}
\usepackage{url}

\usepackage{xcolor, soul}
\sethlcolor{yellow}

\begin{document}
\fancyhead[LO]{Running Title for Header}

\title{VIRL: Volume-Informed Representation Learning towards Few-shot Manufacturability Estimation

}
\author{
  Yu-hsuan Chen,  Jonathan Cagan,  Levent Burak Kara \\
  Department of Mechanical Engineering \\
  Carnegie Mellon University \\
  Pittsburgh, Pennsylvania, 15213, USA\\
  \texttt{\{yuhsuan2, cagan, lkara\}@andrew.cmu.edu} \\
}

\maketitle

\label{sec:abstract}
\begin{abstract}
Designing for manufacturing poses significant challenges in part due to the computation bottleneck of Computer-Aided Manufacturing (CAM) simulations. Although deep learning as an alternative offers fast inference, its performance is dependently bounded by the need for abundant training data. Representation learning, particularly through pre-training, offers promise for few-shot learning, aiding in manufacturability tasks where data can be limited. This work introduces VIRL, a Volume-Informed Representation Learning approach to pre-train a 3D geometric encoder. The pretrained model is evaluated across four manufacturability indicators obtained from CAM simulations: subtractive machining (SM) time, additive manufacturing (AM) time, residual von Mises stress, and blade collisions during Laser Power Bed Fusion process. Across all case studies, the model pre-trained by VIRL shows substantial enhancements on demonstrating improved generalizability with limited data and superior performance with larger datasets. Regarding deployment strategy, case-specific phenomenon exists where finetuning VIRL-pretrained models adversely affects AM tasks with limited data but benefits SM time prediction. Moreover, the efficacy of Low-rank adaptation (LoRA), which balances between probing and finetuning, is explored. LoRA shows stable performance akin to probing with limited data, while achieving a higher upper bound than probing as data size increases, without the computational costs of finetuning. Furthermore, static normalization of manufacturing indicators consistently performs well across tasks, while dynamic normalization enhances performance when a reliable task dependent input is available.
\end{abstract}


\section{Introduction}
\label{sec:intro}
Almost all designs benefit from Computer-Aided Design (CAD) and especially, Computer-Aided Manufacturing (CAM). Understanding how to translate a design into a manufacturable part is paramount in successfully and efficiently transforming a digital prototype into a real-world object. Knowing characteristics, including the time required to manufacture a part, not only aids in improving the design itself but also holds significant implications for supply chain scheduling, cost estimation, and the selection of optimal machining methods \cite{Saric2016, Ning2020}. Additionally, in deposition processes such as casting and additive manufacturing, where components are shaped from high-temperature liquefied material, optimizing part geometry is crucial to prevent undesirable material displacement and residual stress. These factors can lead to deformities, porosity, cracks, or even structural failures. Moreover, in laser power bed fusion (LPBF), excessive part expansion may lead to recoater blade interference, directly jeopardizing the machinery. Being able to design around these manufacturability factors is crucial, which traditionally require prohibitive computer simulations. 

Accurately estimating manufacturability information not only requires laborious simulation setups but also involves significant time investments. In plastic additive manufacturing (AM), a slicer software is typically employed to analyze the part’s geometry. This process involves multiple steps of generating the supports, slicing the model into thin horizontal layers, and then planning the toolpath that the nozzle or laser will follow to deposit material or cure resin \cite{Bi2022}. Similarly, in metal AM, determining residual stresses and thermal failures involves voxelizing the part and conducting finite element analysis (FEA) to simulate thermal and mechanical phenomena \cite{Michaleris2006, Heigel2015, Goldak1984}. Additionally, in CNC machining, minimizing machining time requires intricate knowledge in intelligently selecting tool bits and identifying optimal tool paths and orientations to maximize material removal rate \cite{Armillotta2021}. Due to the complexity of strategy planning, open-source algorithms for end-to-end toolpath generation in CNC machining are currently unavailable, further complicating and prolonging the process. The bottleneck of prohibitive times for manufacturability simulations and the preparation time required exacerbate the challenges in the iterative design process, hindering real-time modifications and adjustments. Overcoming this barrier is crucial for accelerating the design iteration loop, enabling designers to swiftly estimate these manufacturability indicators in the early stage of design and utilize them to optimize the part’s geometry. 

Recently, data-driven methods have emerged as a promising avenue for addressing the issue of lengthy wait times. Leveraging insights gleaned from large datasets, deep learning approaches demonstrate remarkable generalizability in accurately predicting various design-centric properties directly from CAD models \cite{Williams2019, Dering2017}. These predictions span a wide spectrum, including physics-based information such as von Mises stress \cite{Ferguson2024, Nie2020}, temperature \cite{Ferguson2024}, manufacturing cost \cite{Ning2020, Yoo2021}, build time \cite{Sun2022, Chien2023}, AM-central quantities \cite{Williams2019}, and generic product’s functions \cite{Dering2017}. However, a fundamental challenge persists in gathering CAD data. The multitude of design attributes and features presents a daunting task, with limited bandwidth to generate comprehensive datasets for every conceivable case. Moreover, previous efforts often overlook design quantities derived from time-consuming simulations. In essence, while data-driven approaches hold promise in saving time during deployment, the substantial time investment required to gather large datasets paradoxically undermines their intended purpose. 

To combat these shortcomings of fully supervised learning, self-supervised learning has emerged as a promising methodology, where a model autonomously learns to represent data by encoding inherent aspects of the data itself, devoid of supervision from external labels \cite{Liu2023}. This approach has exhibited remarkable efficacy in tasks such as generative AI and in obtaining valuable latent representations for subsequent tasks across diverse domains, spanning text, image, and audio processing. In the realm of Computer-Aided Design (CAD), self-supervised learning has thus far been predominantly employed for tasks such as part classification and surface segmentation \cite{Saric2016, Ning2020}. However, manufacturability indicators in CAD are typically scalar values, necessitating a more challenging regression formulation. Moreover, since prior self-supervised learning tasks have primarily focused on surface-oriented objectives, their direct applicability to manufacturability-related downstream tasks may be limited. Given the abundance of large yet unlabeled datasets in CAD, a natural question arises: Can self-supervised learning be effectively leveraged with an encoder designed to process CAD-native formats as input, enabling it to infer manufacturability indicators even in scenarios with limited labeled data? 

This research, therefore, introduces Volume-Informed Representation Learning (VIRL), a novel pre-training task centered on volumetric mapping, utilizing an encoder-decoder architecture. This approach draws inspiration from practical considerations: subtractive machining involves removing all accessible and unwanted materials from a starting block to create the target part, while additive manufacturing requires the injection head or laser to traverse through every point of the part for infilling. Therefore, equipping the encoder with volumetric data during pre-training is intuitively advantageous for executing such spatial planning tasks. Following self-supervised pre-training, four manufacturability case studies are curated via CAM simulations, involving the prediction of subtractive machining (SM) time, additive manufacturing (AM) print time, maximum residual stresses, and blade collision hazard in LPBF. VIRL-pretrained model’s downstream performance is compared to surface rendering, previously regarded as the state-of-the-art pretraining \cite{Jones2023}, as well as training from scratch. Notably, the comparison involves the same model architecture, with and without pretraining, showcasing the impact of VIRL on manufacturability-oriented downstream tasks.

The main contributions of this work are as follows:
\begin{enumerate}
    \item A novel self-supervised representation learning method specifically tailored to few-shot learning of manufacturability properties from CAD’s native format, utilizing boundary representation. 
    \item A comprehensive dataset comprising over 20,000 mechanical parts, labeled with four key manufacturability indicators, providing a benchmark for future evaluation. 
    \item Discovery of the following insights on the downstream performance of VIRL-pretrained model:
    \begin{enumerate}
        \item AM tasks benefit from probing with multi-layer perceptron (MLP) with limited labeled data, while SM task performs better with finetuning.
        \item Low Rank Adaptation (LoRA) \cite{Hu2021} offers computational savings with a higher upper bound compared to probing with MLP.
        \item Normalizing manufacturability regression tasks statically without incorporating heuristic estimation yields robust performance, while dynamic normalization is advantageous when reliable task-dependent input is available.
    \end{enumerate}
\end{enumerate}

\section{Related Work}
\label{sec:related}
This research aims to estimate manufacturability indicators from CAD parts with limited data, employing self- supervised representation learning. The related work section begins by surveying a spectrum of heuristic approaches employed to estimate manufacturability quantities effectively. Subsequently, it delves into existing methodologies for predicting physical quantities through data-driven techniques. Furthermore, it outlines common pre-training methodologies and highlights finetuning strategies. Lastly, it explores existing research endeavors concentrated on learning from vectorized data, particularly employing graph neural networks. 

\textbf{CAD quantity estimation} spans various methods, encompassing both heuristic and data-driven approaches, for estimating build time, manufacturing cost, stress, temperature, and collision hazard. AM time estimation theoretically can be categorized into two types: parametric, and analytical \cite{Zhang2015}. Parametric approaches consider a few parameters (usually volume, height and surface area) and adopt either regression or empirically learned equations to estimate \cite{Hollis2001}. Because of the simplicity of such mathematical model, the prediction accuracy is often low. On the other hand, analytical approaches adopted by commercial slicing software \cite{Chen1996} considers 180-200 factors including not only part information (e.g., geometries, location, and orientation) but also kinematic factors (laser or nozzle velocity \cite{Chen1996, Campbell2008}), driving factors for tool path (e.g., slicing strategy, contouring area, and hatching length \cite{Giannatsis2001, Campbell2008, Zhang2014}), and support structure \cite{Alexander1998}. While the high accuracy is guaranteed, such model’s high complexity results in lengthy runtime and therefore, becomes disadvantageous to early stage design where on-the-fly inference is essential. For SM, a similar dilemma can also be found in subtractive machining time and cost estimation \cite{Armillotta2021}. On one hand, considering the whole machining process as a sequence of cutting operations on one or more machining tools \cite{Tanner1996, Jha1996, Maropoulos2000, dbenarieh} leads to most accurate results. On the other hand, this analytical method demands human labor to set up CAM simulation and determine operations required for cutting, which is tedious. On the other hand, parametric methods given subtracted volume and other geometry indicator properties using simple regression models find little use except for single machining processes \cite{Creese1992, Swift2013}. Feature-based methods provide a new perspective of estimating the machining time as the summation of contributions from individual surfaces, and assign appropriate removal rate depending on material, surface feature type and parameters. For prismatic parts, Yang and Lin \cite{yang1997} estimate drilling and milling times from removal rates with feature extraction from CAD models. Shehab and Abdalla \cite{Shehab2001} developed a similar approach based on the material removal volume and specified the surface roughness of each feature. Zhang et al. \cite{Zhang2014} proposed a model to estimate machining time based on material removal rates and manufacturing features. Despite the effort, the lack of automation in identifying surface types and their corresponding machining-central features persists to impede rapid inference for CAD designers.

Another emphasis is on the thermal effects that cause AM failure, where the computation cost is even higher. As for physical quantities that relate to potential manufacturing failure modes, current solution turns to numerical approaches employing FEA \cite{Gouge2015, Heigel2015, Vastola2016, Zhang2017}. Part-scale AM process simulation to compute residual stress and distortion is a long time-scale problem involving transient heat transfer, non-linear mechanical deformation, and addition of large amount of materials \cite{Cheng2016, Vastola2016, Mukherjee2017}. Depending on the material under consideration, either a fully coupled or decoupled thermomechanical analysis can be employed. For example, in a decoupled analysis, the thermal analysis is first performed to acquire the temperature distribution at each time step, followed by assigning the obtained thermal load as temperature field at the corresponding step in the mechanical analysis \cite{Denlinger2014, Michaleris2014}. Obviously, the entire process simulation becomes increasingly time-consuming as large amounts of materials are being deposited, and as a result, the size of the modeled part or the number of depositing layers is severely limited. The required simulation time can easily range from several hours to days, or even weeks \cite{Liang2020}. While more efficient algorithms adapted from inherent strain method and applied plastic strain have been proposed and tested \cite{Michaleris2006, Wang2008}, and are being used in commercially, the reality remains that early-stage designers still face significant delays when running simulations during the design loop. Despite these advancements, the current simulation times are improved but still fall short of meeting the demand for real-time feedback, posing a substantial challenge for designers. 


\textbf{Data-driven methods} leveraging deep neural networks for predicting quantities have emerged as promising alternatives for alleviating the computational burden of intensive simulations. These methods offer the capability to directly infer information from geometry, thereby relieving researchers from the need to empirically determine the parameters of each manufacturing method. Moreover, they exhibit the advantage of rapid deployment times. By training across a diverse array of examples, data-driven methods can effectively learn physical phenomena and predict quantities like cost and build time, culminating in the creation of surrogate models capable of accurately predicting results that conventionally come out of simulation \cite{Nourbakhsh2018, Khadilkar2019, Nie2020, barmada2021deep, Jiang2021}.

Numerous studies have concentrated on predicting both the cost and time associated with additive \cite{luca2011, Williams2019, Oh2021} and subtractive \cite{hjahan2001, Zhu2004, yang2007, Ning2020, Yoo2021, Atia2022, Sun2022} manufacturing processes using neural networks. For instance, Ning et al. \cite{Ning2020} take voxelized geometries as input and regress on subtractive manufacturing cost, while Yoo and Kang \cite{Yoo2021} further demonstrate such system’s visual explaniability using gradCAM \cite{selvaraju2017grad}. Notably, Sun et al. \cite{Sun2022} underscore the importance of training with ample data, which tends to yield improved results. Furthermore, it has been demonstrated that with a sufficient dataset, convolutional neural networks (CNNs) and graph neural networks (GNNs) can accurately predict even more complex scalar stress fields \cite{Nie2020, Jiang2021, Ferguson2024}. Specifically, Nie et al. \cite{Nie2020} and Jiang et al. \cite{Jiang2021} utilize pixelated image data to predict stress on 2D structures, while Ferguson et al. \cite{Ferguson2024} use interpolated multi-resolution convolutional neural networks that can operate on arbitrary meshes and achieve higher fidelity results. 

However, it is crucial to acknowledge the drawbacks inherent in this approach. For example, employing CNNs necessitates a non-trivial step of voxelization from CAD-native modalities, which incurs additional computational costs. Moreover, these methods require a substantial number of accurately labeled examples derived from simulations, which can be time-consuming to obtain. 

\textbf{Representation learning}, particularly through self-supervised learning, offers a powerful paradigm that circumvents the need for labeled data, rendering it applicable across a wide range of tasks and modalities. In image processing, self-supervised representation learning has enabled convolutional neural networks (CNNs) and transformer models to achieve faster and sometimes superior classification accuracy through various techniques, including inpainting \cite{Pathak2016, He2022}, rotation \cite{Gidaris2018}, and contrastive learning \cite{Chen2020}. These methods have not only proven that self-supervised learning converges on downstream tasks faster with on-par or even superior performance than training from scratch, but also demonstrated few-shot learning capabilities, where task-related patterns can be learned and generalized by training with very few samples. Similarly, in natural language processing, masked language models, which are trained to predict masked tokens within a sequence based on contextual cues, have attained state-of-the-art performance on benchmarking tasks such as SQuAD, GLUE, and MNLI \cite{Devlin2019, Liu2019}. Another prevalent approach involves pre-training models for next-word prediction, leading to breakthroughs in generative performance, as evidenced by advancements in models like ChatGPT \cite{Openai2018, Radford2019, Brown2020}. 

This form of representation learning has demonstrated its efficacy in leveraging all available data points from source tasks to facilitate few-shot learning on target tasks, both empirically \cite{Brown2020} and theoretically \cite{Du2021}. To adapt pretrained models onto donwstream tasks, the most common strategies are linear probing and finetuning. Probing utilizes the pre-trained model as a fixed feature extractor, with its weights remaining frozen during the downstream task, while only the task-specific layers were trained. Conversely, fine-tuning entailed fine-tuning the entire pre-trained model, including its earlier layers, on the new task-specific data, allowing the model to adapt its learned representations to the characteristics of the new dataset. While in most cases fine-tuning exudes superior performance over probing, it has the downside of high computational cost especially when the model is at millions or even billions scale \cite{Bommasani2021}. To efficiently fine-tune models on downstream tasks while surpassing the performance of probing methods, researchers have explored parameter-efficient tuning techniques \cite{Hu2021, Lester2021, He2022}. Among these methods, low-rank adaptation (LoRA) has emerged as a popular choice, capable of reducing the number of trainable parameters by up to 10,000 times and the GPU memory requirements by threefold, all while maintaining or surpassing the quality of models fine-tuned conventionally, particularly on large language models \cite{Hu2021}. This technique involves attaching low-dimensional adaptation matrices to desired layers, thereby directing features towards the target downstream task without distorting the pre-learned weights. Notably, low-rank adaptation has been successfully applied to graph neural networks (GNNs) \cite{Li2024} and convolutional neural networks (CNNs) \cite{Zhong2023}. 

However, a key consideration arises when transitioning these techniques to design workflows, particularly when working with vectorized data from Computer-Aided Design (CAD) or non-Euclidean datasets: do these methods remain effective in such contexts? 

\textbf{Learning from vectorized representations} typically involves constructing hierarchical Graph Neural Networks (GNNs), a technique that has been extensively studied in both 2D and 3D domains. In 2D vector graphics (SVGs), GNNs have been applied to tasks such as manufacturing classification \cite{Xie2022}, engineering drawing segmentation \cite{Zhang2023}, and stylized feature extraction \cite{Chen2024}. In the realm of 3D representations, researchers have primarily conceptualized Boundary Representation (BRep) models as hierarchical graphs \cite{Jones2021, pradeep2021, Lambourne2021, Jones2023}. In this framework, a CAD part is envisioned as a hierarchical graph structure, wherein mutually connected surfaces form the foundational nodes. Each surface within this hierarchy possesses its distinct features, while their associated curves are interconnected. These curves, in turn, harbor their own unique characteristics, and their vertices are likewise characterized by features. Notably, UV-Net \cite{pradeep2021} proposed a method to parameterize each geometry into the U and V parameter domains of curves and surfaces, demonstrating superior performance in part classification and surface segmentation tasks. Additionally, UV-Net conducted self-supervised learning experiments using a Siamese network-based contrastive learning approach and achieved successful shape retrieval outcomes. 

An improved self-supervised learning approach taken by researchers involves learning CAD’s implicit representation by mapping pre-defined graph features to reconstruct the surface sign distance field on the UV domain \cite{Jones2023}. Essentially, this approach entails mapping explicit surface features to an implicit boundary representation, yielding notable improvements in few-shot learning performance and achieving state-of-the-art results in part classification and segmentation tasks. 

However, a significant research gap persists: previous studies have not focused on manufacturing-related tasks. Since the nature of regression and physics is very different from classification and segmentation, it is questionable that pretraining techniques designed for prior tasks can transfer well onto manufacturability domains. Furthermore, the absence of a predefined task poses challenges, as running manufacturability simulations is both time-consuming and difficult to automate. Consequently, key questions arise: is it feasible to construct a manufacturability dataset, and what pre-training scheme would best facilitate few-shot learning for manufacturability-related tasks?

\section{Methods}
\label{sec:methods}
This research aims to estimate manufacturability indicators from CAD parts with limited data, utilizing self-supervised representation learning. This section presents the proposed self-supervised pre-training framework and outlines the setup for the four manufacturability case studies derived from simulation to evaluate the effectiveness of the pretrained model. 
\subsection{Proposed architecture}

\begin{figure}[H]
    \centering
    \includegraphics[width=\textwidth]{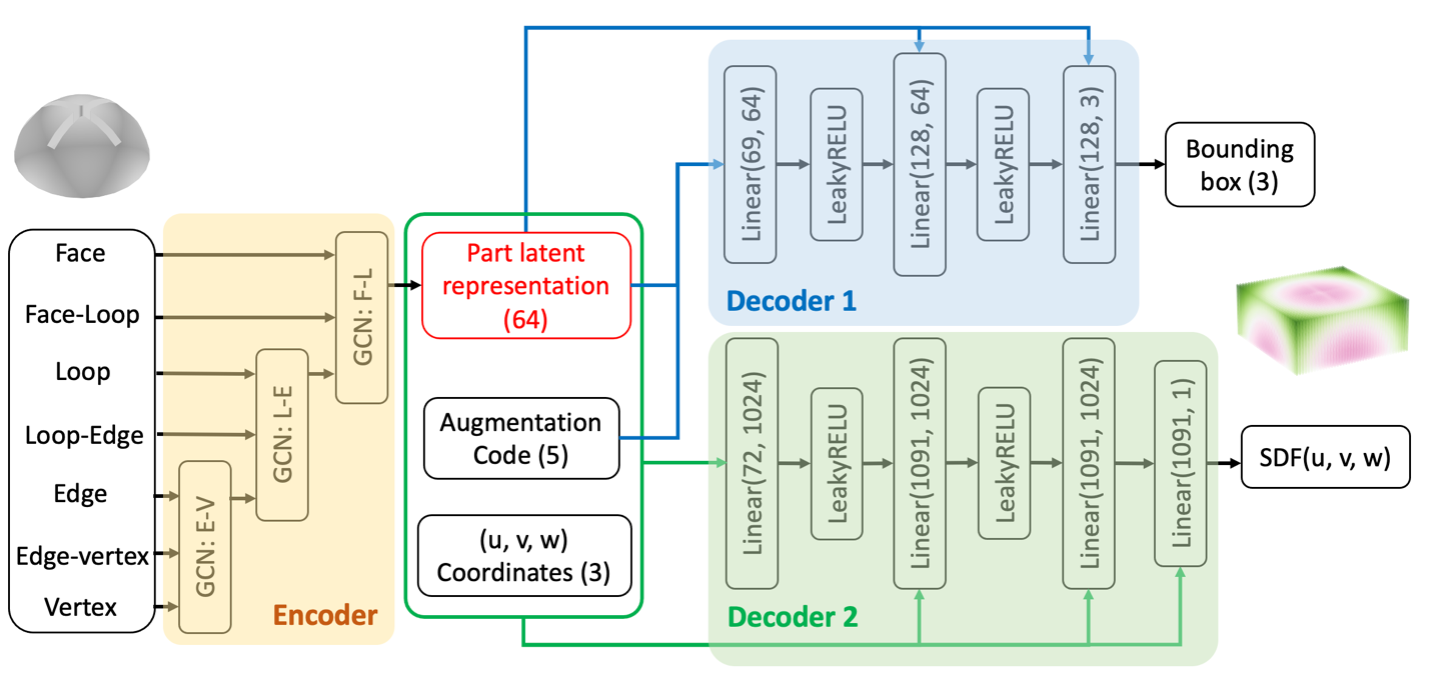} 
    \caption{Schematic diagram of the VIRL’s architecture. The numbers in parentheses represent the dimension of latent vectors and linear layers.}
    \label{fig:1}
\end{figure}
\begin{table}[H]
    \centering
    \caption{Hyperparameters of pre-training with VIRL}
    \renewcommand{\arraystretch}{1.2}
    \begin{tabular}{c|c|c|c|c|c|c|c}
    Optimizer & 
    \begin{tabular}[c]{@{}c@{}}Learning\\ 
        rate\end{tabular} & 
    Scheduler & Loss & Encoder & Decoder 1 & Decoder 2 & 
    \begin{tabular}[c]{@{}c@{}}Batch\\ 
        size\end{tabular}\\

        \hline
        Adam & 
        \begin{tabular}[c]{@{}c@{}}$10^{-4}$\\
        $\rightarrow 10^{-6}$\end{tabular} & 
        \begin{tabular}[c]{@{}c@{}}cosine\\ 
        annealing\end{tabular} & 
        MSE & 
        \begin{tabular}[c]{@{}c@{}}3-tier GCN\\ 
        6.4 M params\end{tabular} & 
        \begin{tabular}[c]{@{}c@{}}2-layer MLP\\ 
        13K params\end{tabular} & 
        \begin{tabular}[c]{@{}c@{}}4-layer MLP\\
        2.3 M params\end{tabular} & 
        64

    \end{tabular}

    \label{tab:1}
\end{table}

Inspired by the spatial organization found in manufacturing simulation, this study introduces Volume-Informed Representation Learning (VIRL), a novel self-supervised learning framework that combines boundary representation (BRep) encoding with volumetric comprehension. At the heart of this methodology lies a distinctive pretraining task centered on volumetric mapping, facilitated by an encoder-decoder architecture. The encoder processes BRep inputs to generate a latent vector, which acts as the implicit representation of the component. Considering the multifaceted nature of CAD parts, involving various faces, loops, edges, and vertices, the encoder employs a triple hierarchy Graph Convolutional Neural Network (GCN) architecture, as depicted in Fig. \ref{fig:1}.

To facilitate pretraining with volumetric information, two multi-layer perceptron (MLP) decoders are appended to the latent output. Firstly, Decoder 1 is tailored to regress the bounding box value, providing the model with insights into the scale and proportions of the domain. The effectiveness of incorporating bounding box information during training has been demonstrated in previous studies \cite{Chen2024}. Secondly, Decoder 2 aims to reconstruct the geometry in a point-wise spatial field modality from the latent representation, leveraging its demonstrated ability to achieve exceptional shape representation and completion for complex topologies \cite{Park2019}.

Decoder 2 utilizes both the latent representation and UVW (normalized XYZ) grid input to predict the sign distance field (SDF) value from the closest boundary, corresponding to the given UVW coordinate. Each set of (u, v, w)  coordinate in the UVW grid consists of three scalar values ranging from 0 to 1, representing a unit cube scaled from the bounding box of the part. The choice of using normalized UVW spectrum instead of XYZ as input is due to XYZ being learned from Decoder 1, offering an equivalent yet more simplistic alternative.

To ensure efficient sign distance retrieval at any arbitrary (u, v, w) coordinate during pretraining, the SDF for each part is uniformly pre-computed with 40 values per UVW grid axis. This allows for the utilization of trilinear interpolation, enabling swift access to sign distance values. Additionally, to bias the sample towards the 0-level set based on empirical evidence that will benefit detailed reconstruction \cite{Park2019, Jones2023}, 40\% of sampling points are selected within a distance of 0.01 from the boundary and the rest are randomly sampled within the bounding box. The pretraining hyperparameters are listed in Table \ref{tab:1}. 

\begin{figure}[H]
    \centering
    \includegraphics[width=\textwidth]{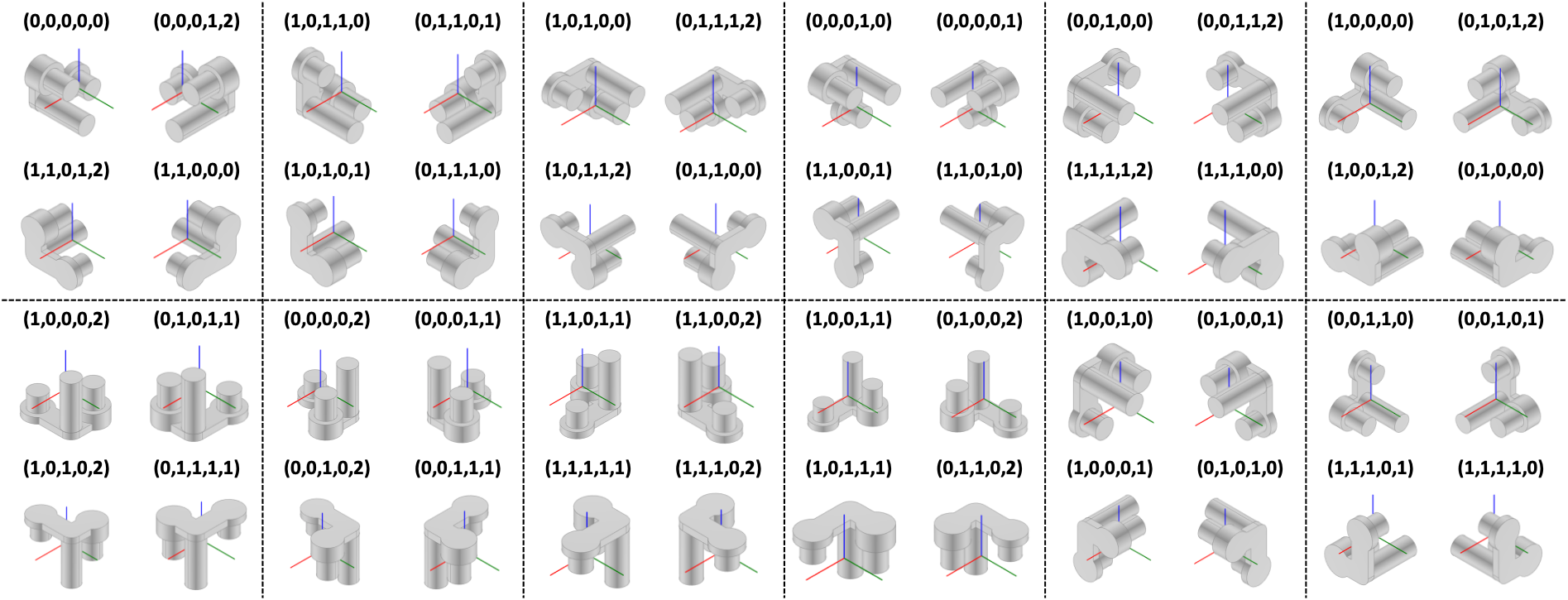} 
    \caption{48 variations of a part being transposed and rotated and their corresponding 5-variable augmentation code.}
    \label{fig:2}
\end{figure}

\begin{figure}[H]
    \centering
    \includegraphics[width=\textwidth]{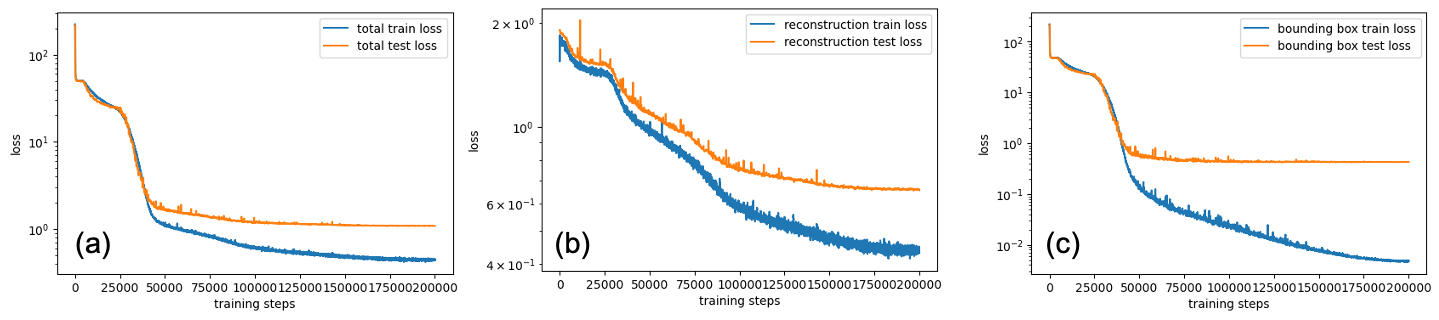} 
    \caption{The loss curves during pretraining. (a) total loss (Decoder1 + Decoder2) (b) reconstruction loss (Decoder2), and (c) bounding box loss (Decoder1).}
    \label{fig:3}
\end{figure}

\begin{figure}[H]
    \centering
    \includegraphics[width=\textwidth]{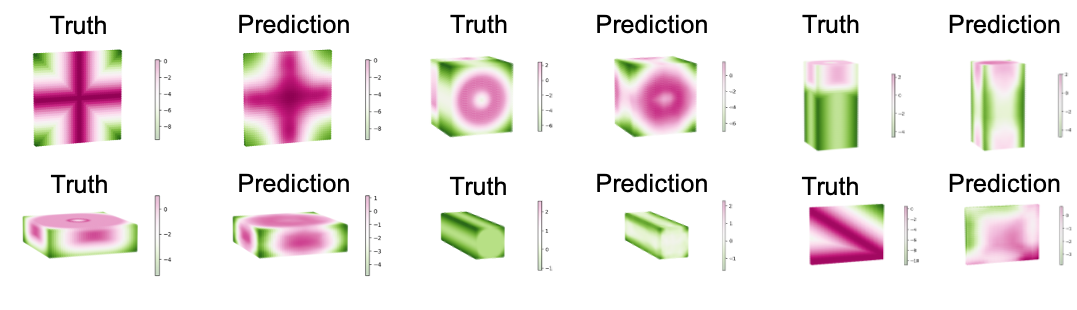} 
    \caption{SDF reconstruction results in the test set.}
    \label{fig:4}
\end{figure}

In the decoding phase, augmentation techniques such as random flipping and axis swapping are applied, incorporating a 5-variable vector to represent the augmentation code, expanding the dataset by 48 times. An example of 48 distinct geometries with their corresponding augmentation code is illustrated in Fig. \ref{fig:2}. Specifically, during decoding, the latent code for different augmentation code remains constant, while the ground truth is modified, yielding 48 different outcomes of each part’s corresponding objective bounding box and sign distance field. This augmentation is introduced at the intermediate stage to bolster dataset diversity without the need to regenerate graph input by the CAD kernel, which is time-consuming. 

During pre-training, back propagation is performed on a combined loss of the bounding box regression and sign distance field reconstruction. The loss curve during pretraining, as depicted in Fig. \ref{fig:3}, demonstrates that bounding box loss dominates initially and plateaus after 50000 iterations, while the reconstruction loss continues to descend. Following pretraining, the model with the highest testing accuracy is selected, and its SDF reconstruction from Decoder 2 on a uniform UVW grid scaled with Decoder 1’s bounding box prediction of selected testing data is depicted in Fig. \ref{fig:4}. The results demonstrate a reasonable reconstruction of the majority of SDF fields, highlighting the model’s generalizability. However, some limitations are observed in more complex geometries, suggesting areas for further improvement. 

\subsection{Case Studies Formulation}

\begin{table}[H]
    \renewcommand{\arraystretch}{1.5}
    \setlength{\tabcolsep}{3pt}
    \centering
    \caption{List of 4 manufacturability case studies’ properties.}
    \begin{tabular}{c|c|c|c|c|c|c}
    Quantity &
    \begin{tabular}[c]{@{}c@{}}Manufacture\\ 
    type\end{tabular} & 
    \begin{tabular}[c]{@{}c@{}}Material\\ 
    Usage\end{tabular} & 
    \begin{tabular}[c]{@{}c@{}}Material\\ 
    type\end{tabular} & 
    Machine &
    Software &
    Pre-processing \\
        \hline
        Residual stress &
        Metallic LPBF & 
        Additive &
        Inc625 40$\mu$m &
        AM250 &
        Netfabb &
        NA \\
        
        \hline
        Blade Collision &
        Metallic LPBF &
        Additive &
        Inc625 40$\mu$m &
        AM250 &
        Netfabb &
        NA \\
        
        \hline
        Printing time &
        \begin{tabular}[c]{@{}c@{}}Plastic 3D\\ 
        printing\end{tabular} & 
        Additive &
        Generic PLA &
        \begin{tabular}[c]{@{}c@{}}Prusa i3\\ 
        Mk3\end{tabular} & 
        CURA &
        \begin{tabular}[c]{@{}c@{}c@{}}Generate\\ 
        adhesion and\\
        support structures\end{tabular} \\
        
        \hline
        Cutting time &
        \begin{tabular}[c]{@{}c@{}}CNC\\ 
        machining\end{tabular} & 
        Subtractive &
        Not specified &
        Not specified & 
        Siemens NX &
        \begin{tabular}[c]{@{}c@{}c@{}}Setup\\ 
        determination\end{tabular} \\

    \end{tabular}
    
    \label{tab:2}
\end{table}

\begin{figure}[H]
    \centering
    \includegraphics[width=\textwidth]{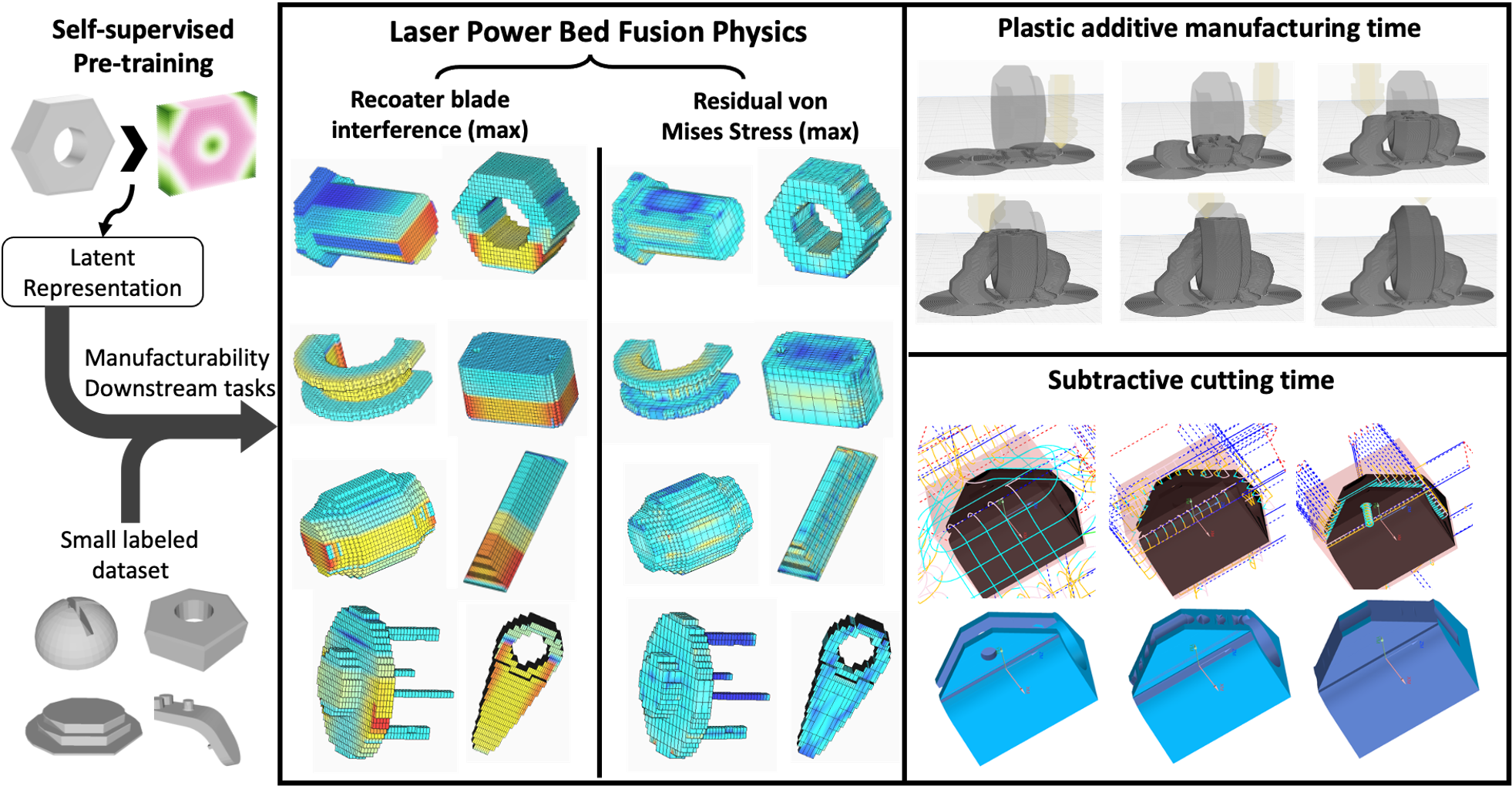} 
    \caption{Downstream regression pipeline. After the encoder is pre-trained with VIRL, the latent representation is used to regress on four manufacturability cases studies curated from simulation.}
    \label{fig:5}
\end{figure}

To assess the performance of VIRL-pretrained model on manufacturability estimation, simulations on mechanical parts sourced from Fusion360 segmentation dataset \cite{Lambourne2021} are conducted. The four cases studies and key properties are listed in Table \ref{tab:2}, while the downstream regression pipeline and simulation examples are visualized in Fig. \ref{fig:5}. For each case study, automated, generic pipelines are established for preprocessing and importing the part into simulation. Subsequently, by gathering simulation data, one could evaluate the downstream regression performance to determine the effectiveness of VIRL, where the size of the labeled dataset is varied to evaluate the model’s performance across both limited data (100 samples) and abundant data (10000 samples). 

Firstly, understanding the significance of physics-based manufacturability information, Netfabb simulations on the Laser Powder Bed Fusion (LPBF) manufacturing process is conducted. LPBF can lead to thermal expansion, resulting in two potential issues: upward bending of the workpiece during the process, which may cause collisions with the blade, and residual stress post-manufacturing, leading to defects like porosity and cracks. To explore these aspects, blade interference hazard and residual stress are selected as key case studies. The simulation utilizes an AM250 machine with Inc625 40um material in Netfabb, employing voxelization and Finite Element Analysis (FEA) for both thermal and mechanical data generation. Subsequently, successful simulation results are extracted and the maximum values of von Mises stress and blade interference hazard in percentage are identified as labels, as these are indicative of the two potential failure modes. 

Regarding Additive Manufacturing (AM) time estimation, it was found that LPBF time was highly correlated with part height due to the rapid scanning of the laser. However, estimating print time becomes more challenging when considering support structure print time, especially when it is not specifically learned in the geometry during pre-training. To explore this, plastic AM slicing simulations are conducted using CURA. CURA performs adhesion, support structure generation, slicing, and toolpath generation, ultimately outputting an accumulative print time including infilling, wall, support structure, adhesion, travelling, retraction, and various resolution and machine parameters such as moving speed, layer height, density, etc. Table \ref{tab:3} lists crucial configuration settings. 

\begin{table}[H]
    \centering
    \caption{List of AM print time configuration settings.}
    \renewcommand{\arraystretch}{1.2}
    \begin{tabular}{c|c|c|c|c|c|c|c|c}
    \begin{tabular}[c]{@{}c@{}}infill\\ 
        density\end{tabular} & 
    \begin{tabular}[c]{@{}c@{}}infill\\ 
        pattern\end{tabular} & 
    \begin{tabular}[c]{@{}c@{}}overhang\\ 
        angle\end{tabular} & 
    \begin{tabular}[c]{@{}c@{}}support\\ 
        structure\end{tabular} &         
    \begin{tabular}[c]{@{}c@{}}layer\\ 
        height\end{tabular} & 
    \begin{tabular}[c]{@{}c@{}}shell\\ 
        thickness\end{tabular} & 
    \begin{tabular}[c]{@{}c@{}}wall\\ 
        count\end{tabular} & 
    \begin{tabular}[c]{@{}c@{}}print\\ 
        speed\end{tabular} & 
    \begin{tabular}[c]{@{}c@{}}build plate\\ 
        adhesion\end{tabular}\\

        \hline
        20\% & grid & 50 & tree &
        \begin{tabular}[c]{@{}c@{}}0.15\\ 
            mm\end{tabular} &
        \begin{tabular}[c]{@{}c@{}}0.8\\ 
            mm\end{tabular} &
        2 &
        \begin{tabular}[c]{@{}c@{}}60\\ 
            mm/s\end{tabular} &
        brim \\
            
    \end{tabular}
    
    \label{tab:3}
\end{table}

\begin{figure}[]
    \centering
    \includegraphics[width=\textwidth]{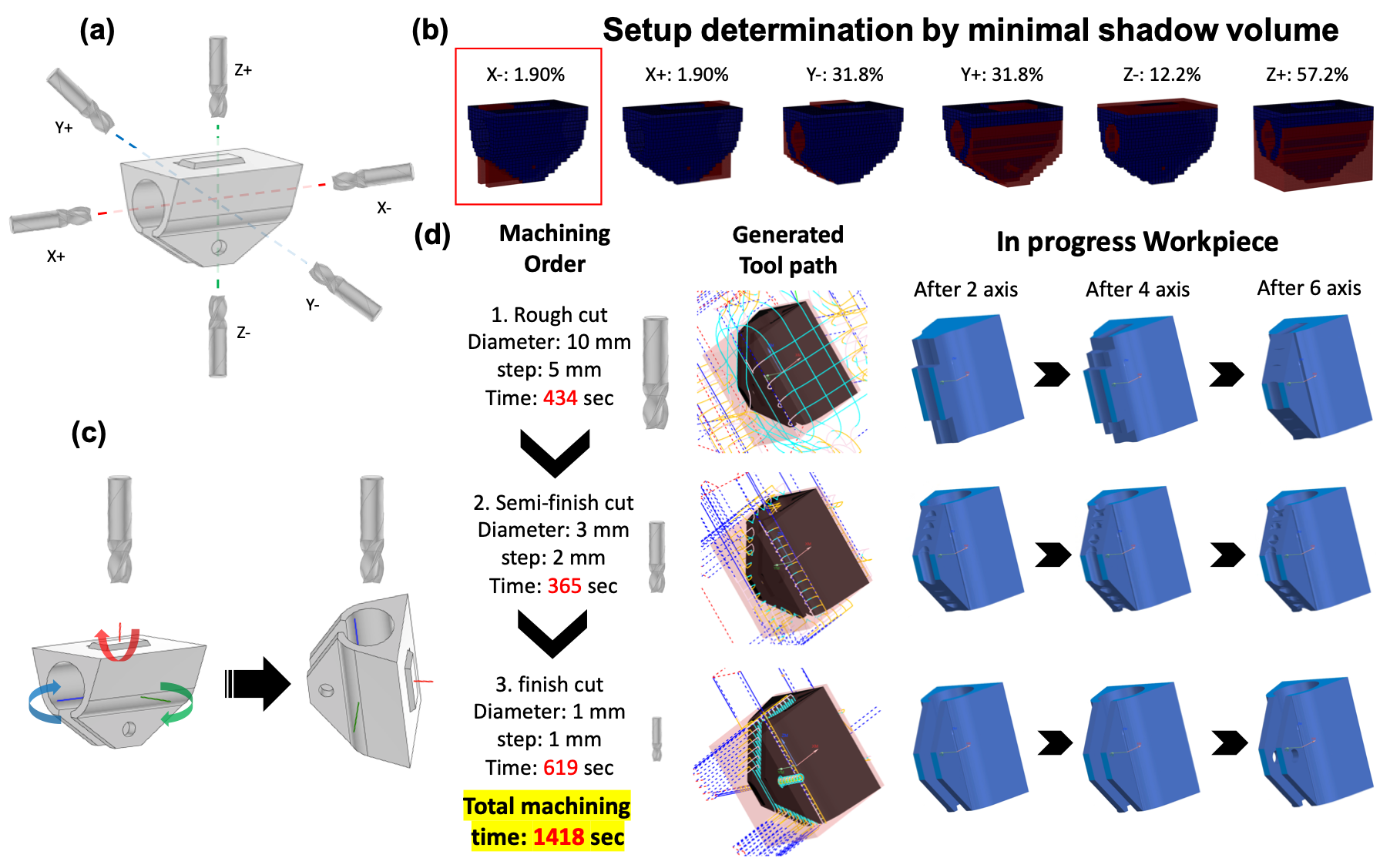} 
    \caption{3+2 axis machining simulation pipeline. (a) The six candidates’ starting orientation. (b) Shadow volume calculation (c) orienting the part to the axis with minimal shadow volume, and (d) three tool bits will sequentially subtract volumes from the stock for all six axes.}
    \label{fig:6}
\end{figure}

In addition, to explore the model’s capability on estimating subtractive machining quantity, a 3+2 axis CNC cutting time case study is conducted. Unlike additive manufacturing, CNC machining presents greater challenges in automation due to the need for strategic part orientation, tool selection, and cutting methodologies. To create a simulation pipeline that balances realism with robustness across the dataset, the following approach is devised, illustrated in Fig. \ref{fig:6}. Firstly, the optimal setup orientation is determined by assessing trapped volume along all six axes \cite{Nelaturi2015, Langelaar2019}. The axis with the least inaccessibility is chosen as the initial cutting orientation, followed by cutting from remaining five axes if the part needs further subtraction from the stock. Additionally, three sequential tool bits are employed to cut the stock, each progressively smaller to simulate roughing, semi-finishing, and finishing processes. As a result, the cutting procedure could potentially involve up to 18 processes in total. Utilizing Siemens NX for CAM simulation, it is observed that the software effectively machined the majority of parts in the Fusion360 segmentation dataset. It is worth noting that if a part can be successfully machined without requiring all 18 processes, the machining process terminates upon completion of the cutting. It is assumed negligible setup switching time, and therefore the summation of all 18 processes constitutes the total machining time. 

\begin{figure}[]
    \centering
    \includegraphics[width=\textwidth]{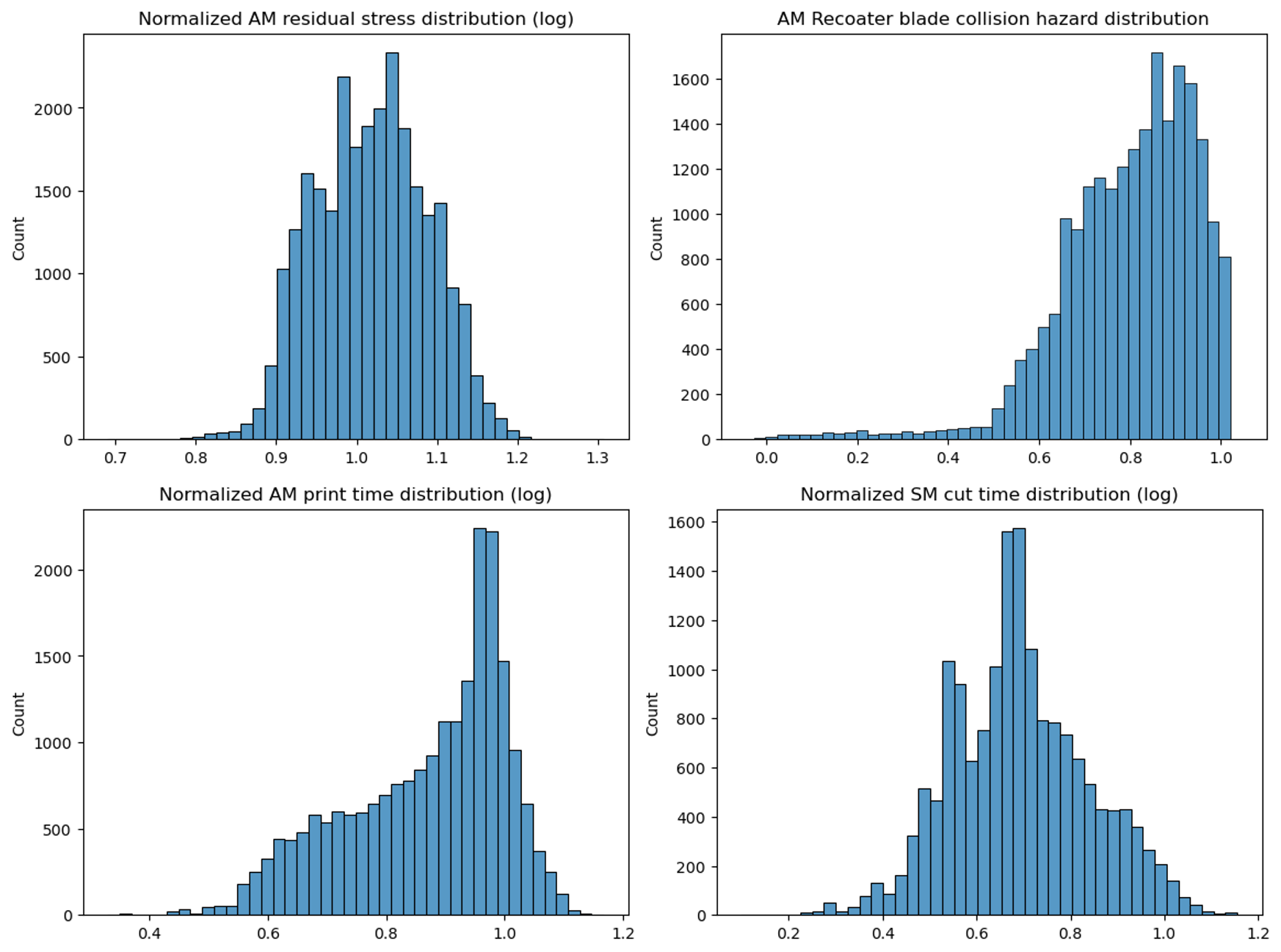} 
    \caption{Data distribution of the four case studies.}
    \label{fig:7}
\end{figure}

Fig. \ref{fig:7} illustrates the distribution of the manufacturability indicators collected from all successful simulations from the four case studies. For subtractive machining (SM) and additive manufacturing (AM) time, as well as AM residual von Mises stress, given their wide-ranging values up to the thousand scale, the distributions are displayed after normalization in the logarithmic domain. This normalization helps alleviate the challenge faced by downstream neural networks in regressing such diverse data. However, blade collision percentage values predominantly fall within the range of 0 to 1, which is already conducive to downstream regression tasks. Hence, these values are directly regressed upon in downstream evaluation. Notably, all four case studies exhibit distributions that closely approximate a Gaussian shape. 

\subsection{Testing Pretrained Model on Downstream Tasks}

After acquiring spatial and volumetric comprehension during VIRL-pretraining, the graph neural network encoder is expected to output latent representation optimal for various manufacturability tasks. Post pretraining, the decoder is discarded, and the encoder’s latent code layer is extended with two additional linear layers tasked with regressing on the four specified manufacturability indicators. To gauge the utility and effectiveness of pretrained models on downstream tasks, experiments are conducted across multiple dimensions. 

VIRL-pretrained model is first compared with two alternatives: training from scratch and the surface rendering method, previously regarded as the state-of-the-art BRep pretraining technique \cite{Jones2023}. Using dimensionality reduction, the latent space is visualized to illustrate how well the input data aligns with manufacturability indicators. Next, the performance of these models is evaluated on datasets of varying sizes, ranging from 50 to 10000 samples. Both probing and fine-tuning methods are deployed to assess performance, with probing revealing stability as a feature extractor and finetuning exploring adaptability to task specific data. Probing methodologies included SVR, MLPs, and LoRA, providing insights into optimal use cases. 

Furthermore, inclusion of task dependent input is investigated in downstream tasks, examining its necessity and methodologies. Specifically, experiments focus on how incorporating parametric estimation affected the regression performance of AM print time and SM cutting time. Moreover, ablation studies are conducted on two hyperparameters, namely hidden layer width and latent code width, to assess the model’s scalability and performance across varying configurations.

\section{Results and Discussion}
\label{sec:results}

\begin{figure}[h]
    \centering
    \includegraphics[width=\textwidth]{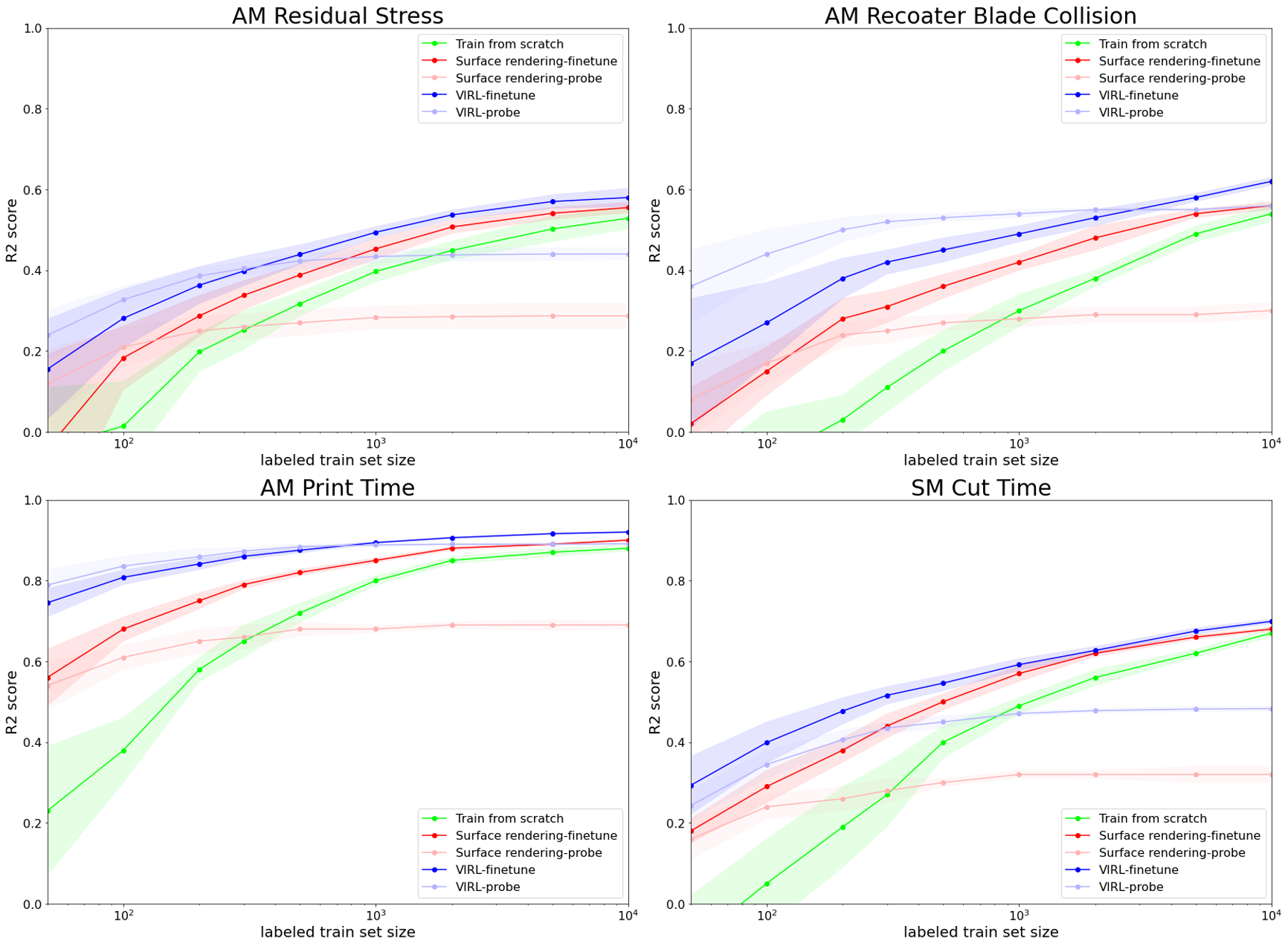} 
    \caption{Regression R2 scores for four manufacturability case studies. Each plot, from left to right, shows models’ performance from few-shot learning (50 samples), to having abundant data (10000 samples). Each data point represents the average of 10 runs, with the displayed bandwidth indicating the standard deviation.}
    \label{fig:8}
\end{figure}

Fig. \ref{fig:8} presents the R2 scores for four manufacturability case studies with VIRL pretraining, training from scratch, and surface rendering pretraining across varying data sizes ranging from 50 labeled data (few-shot) to larger datasets (10,000 samples). Notably, all models operate without task dependent input, relying solely on the latent code output from the encoder. Note that all models have the exact same architecture, so pretraining is the only variable. The downstream training is done by attaching 2 linear layers and regressing with Huber loss. Observations reveal a consistent outperformance of VIRL-pretrained model over both training from scratch and surface rendering pretraining method, demonstrating better performance across few-shot scenarios (50 shots) and achieving a higher upper bound given more data. Although in all case studies, the performance of the 3 models, when being fully updatable, as expected, converge to approximately similar R2 performance when there is ample data (10000 samples), the difference when there is limited data (\textless100 samples) is not negligible. The fact that VIRL pretraining is better than surface rendering pretraining is particularly evident in the probing results. Additionally, specific insights are noted: while AM print time prediction proves relatively straightforward as being the only case study that VIRL’s R2 score is able to surpass 0.9, SM cut time estimation poses greater challenges compared to AM print time. Although VIRL-pretrained model still faces difficulties in learning physics-based values such as stress and blade collision hazard, its performance has significantly improved by generalizing from a smaller number of samples. Interestingly, probing outperforms finetuning in cases where the dataset contains fewer than 1,000 samples, except for SM cut time. This is likely attributed to the task’s surface-oriented nature, where updating the volumetric reconstruction pretrained model from the outset proves beneficial. 

\subsection{Models' Performance on Various Tasks and Dataset Size}

\begin{figure}[]
    \centering
    \includegraphics[width=\textwidth]{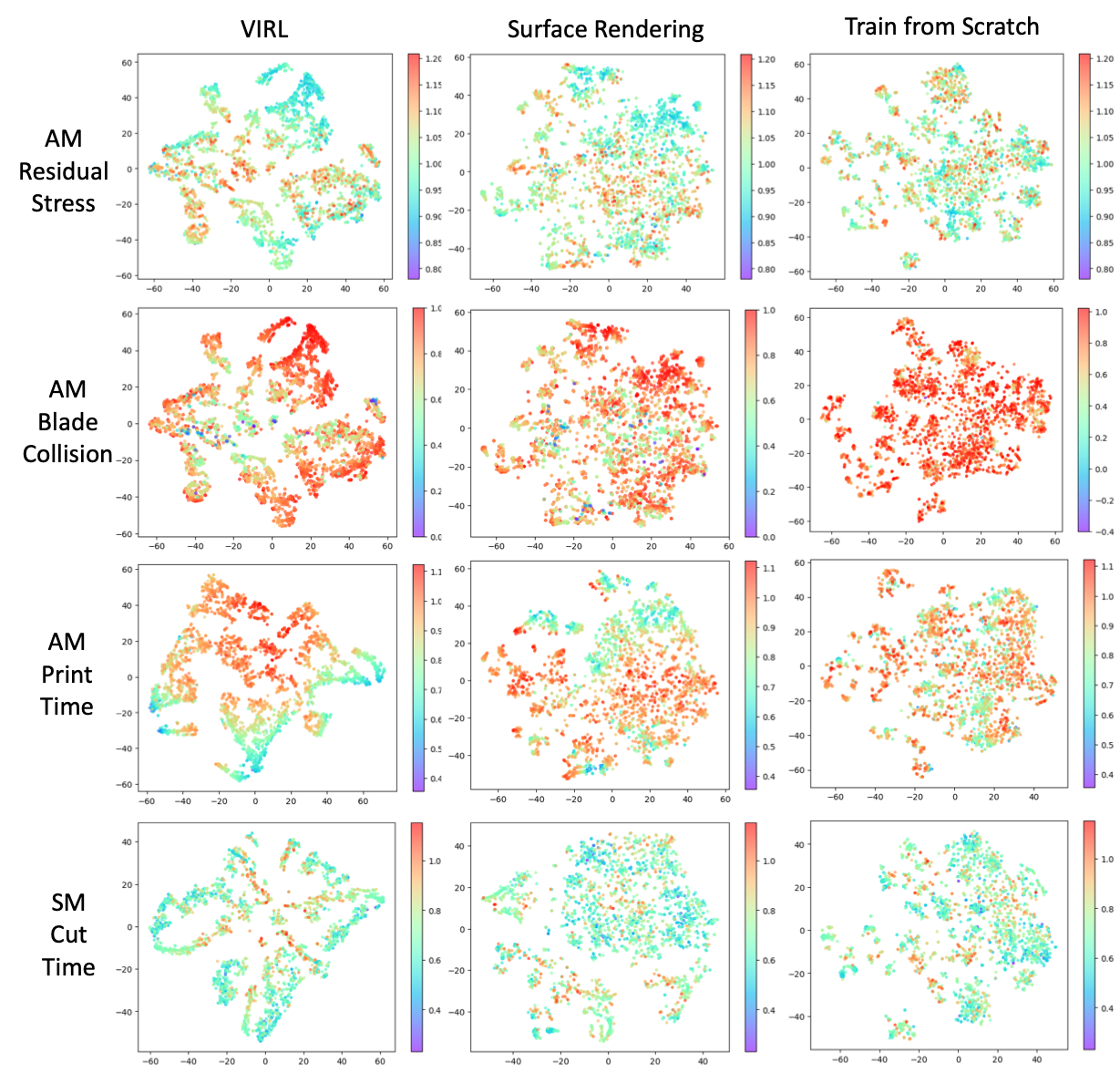} 
    \caption{tSNE latent space visualization. Each point is colored with their corresponding manufacturability indicators.}
    \label{fig:9}
\end{figure}

\subsection{Visualizing Latent Space via Dimension Reduction}
To visually elucidate the efficacy of the VIRL-pretrained model, Fig. \ref{fig:9} presents 2D tSNE plots of latent vectors output from the encoder prior to downstream tasks, with colors indicating the scalar values of all four case studies. A comparison against models trained from scratch and pretrained with surface rendering reveals that the latent code distribution pretrained with VIRL better aligns with the manufacturability values, particularly noticeable in AM print time. This observation offers insight into why VIRL’s downstream performance generally outperforms other methods, attributed to the superior starting latent space. Furthermore, the visualization suggests that AM print time is the least challenging task, while learning physics-based information such as stress and blade collision hazard presents greater difficulty, matching the experimental results in Fig. \ref{fig:8}. 

\begin{figure}[H]
    \centering
    \includegraphics[width=10cm]{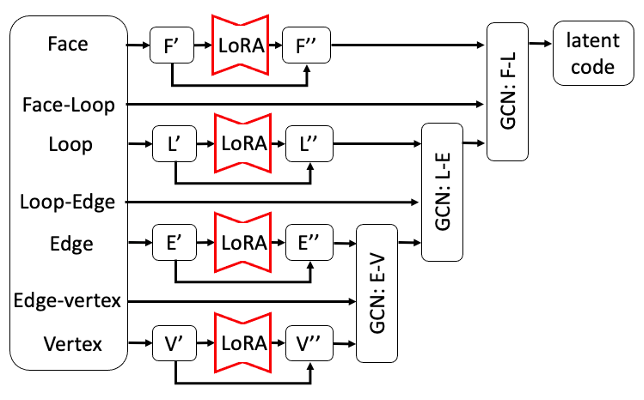} 
    \caption{LoRA results reported are applied to the early layers in the encoder, where it is found to achieve optimal performance.}
    \label{fig:10}
\end{figure}

\begin{table}[H]
\renewcommand{\arraystretch}{1.4}
\centering
\caption{Downstream performance vs different probing and finetuning strategies.}
\begin{tabular}{cc|c|c|c|c|c}
\multicolumn{2}{c|}{\textbf{parameters}} &
  \textbf{18} &
  \textbf{4225} &
  \textbf{4737} &
  \textbf{8321} &
  \textbf{6.4 M} \\ \hline
\multicolumn{1}{c|}{\textbf{task}} &
  \textbf{shots} &
  \textbf{Probe - SVR} &
  \textbf{Probe – MLP} &
  \textbf{LoRA – rank 1} &
  \textbf{LoRA – rank 8} &
  \textbf{Finetune} \\ \hline
\multicolumn{1}{c|}{} &
  100 &
  {\color[HTML]{00ca5b} 0.316±0.032} &
  0.345±0.034 &
  {\color[HTML]{00ca5b} 0.376±0.059} &
  0.396±0.055 &
  {\color[HTML]{cc0000} 0.399±0.051} \\ \cline{2-7} 
\multicolumn{1}{c|}{} &
  1k &
  {\color[HTML]{00ca5b} 0.469±0.005} &
  0.471±0.006 &
  {\color[HTML]{00ca5b} 0.541±0.011} &
  0.582±0.015 &
  {\color[HTML]{cc0000} 0.592±0.015} \\ \cline{2-7} 
\multicolumn{1}{c|}{\multirow{-3}{*}{\begin{tabular}[c]{@{}c@{}}SM\\cut\\time\end{tabular}}} &
  10k &
  {\color[HTML]{cc0000} 0.533±0.002} &
  {\color[HTML]{00ca5b} 0.483±0.006} &
  0.557±0.012 &
  0.602±0.009 &
  {\color[HTML]{cc0000} 0.699±0.006} \\ \hline
\multicolumn{1}{c|}{} &
  100 &
  {\color[HTML]{00ca5b} 0.742±0.049} &
  {\color[HTML]{cc0000} 0.831±0.029} &
  {\color[HTML]{00ca5b} 0.822±0.032} &
  0.818±0.017 &
  0.807±0.028 \\ \cline{2-7} 
\multicolumn{1}{c|}{} &
  1k &
  {\color[HTML]{00ca5b} 0.825±0.010} &
  0.868±0.010 &
  {\color[HTML]{00ca5b} 0.871±0.008} &
  0.870±0.005 &
  {\color[HTML]{cc0000} 0.889±0.003} \\ \cline{2-7} 
\multicolumn{1}{c|}{\multirow{-3}{*}{\begin{tabular}[c]{@{}c@{}}AM\\print\\time\end{tabular}}} &
  10k &
  {\color[HTML]{00ca5b} 0.855±0.004} &
  0.872±0.007 &
  {\color[HTML]{00ca5b} 0.876±0.007} &
  0.875±0.005 &
  {\color[HTML]{cc0000} 0.914±0.002} \\ \hline
\multicolumn{1}{c|}{} &
  100 &
  {\color[HTML]{00ca5b} 0.422±0.029} &
  {\color[HTML]{cc0000} 0.445±0.058} &
  0.414±0.069 &
  0.369±0.055 &
  {\color[HTML]{00ca5b} 0.274±0.096} \\ \cline{2-7} 
\multicolumn{1}{c|}{} &
  1k &
  {\color[HTML]{cc0000} 0.545±0.011} &
  {\color[HTML]{cc0000} 0.545±0.007} &
  {\color[HTML]{cc0000} 0.543±0.009} &
  0.526±0.012 &
  {\color[HTML]{00ca5b} 0.495±0.017} \\ \cline{2-7} 
\multicolumn{1}{c|}{\multirow{-3}{*}{\begin{tabular}[c]{@{}c@{}}AM\\Blade\\collision\end{tabular}}} &
  10k &
  {\color[HTML]{cc0000} 0.595±0.003} &
  0.555±0.004 &
  0.558±0.008 &
  {\color[HTML]{00ca5b} 0.550±0.013} &
  {\color[HTML]{cc0000} 0.618±0.012} \\ \hline
\multicolumn{1}{c|}{} &
  100 &
  {\color[HTML]{00ca5b} 0.251±0.076} &
  0.327±0.034 &
  {\color[HTML]{cc0000} 0.343±0.033} &
  {\color[HTML]{cc0000} 0.323±0.040} &
  0.281±0.072 \\ \cline{2-7} 
\multicolumn{1}{c|}{} &
  1k &
  {\color[HTML]{00ca5b} 0.385±0.027} &
  0.434±0.017 &
  {\color[HTML]{00ca5b} 0.446±0.014} &
  0.444±0.016 &
  {\color[HTML]{cc0000} 0.494±0.014} \\ \cline{2-7} 
\multicolumn{1}{c|}{\multirow{-3}{*}{\begin{tabular}[c]{@{}c@{}}AM\\residual\\stress\end{tabular}}} &
  10k &
  {\color[HTML]{00ca5b} 0.417±0.005} &
  0.440±0.015 &
  {\color[HTML]{00ca5b} 0.454±0.015} &
  0.455±0.014 &
  {\color[HTML]{cc0000} 0.579±0.023} \\\\
\end{tabular}

\label{tab:4}
\end{table}

\subsection{Exploring Probing and Finetuning Alternatives for VIRL-pretrained Model}
To further understand the optimal usage of VIRL-pretrained model, given limited data or computation resources, this section study four deployment strategies, namely probing with Support Vector Regression (SVR), probing with multi-layer perceptron (MLP), Low-rank adaptation (LoRA), and fully finetuning. Regarding LoRA, it is found that applying adaptation at the early layers of the encooder yields the best results, shown in Fig. \ref{fig:10}. To understand the effect of rank width, which is the bottleneck dimension of the appended matrices, rank of 1 and 8 are studied. For SVR, experiments revealed that Radial Basis Function (RBF) tends to achieve the best performance. Each deployment strategy is studied on all four case studies, provided with 100, 1000, and 10000 training samples, shown in Table \ref{tab:4}. 

This experiment yields several findings. First of all, comparing probing with MLP and SVR, while SVR has the advantage of computational efficiency by having very few learnable parameters, SVR underperforms probing with MLP when considering limited data in all four case studies. Secondly, in all three AM case studies, finetuning the entire model tends to yield worse results than probing with MLP and LoRA adaptation, especially with only 100 shots. This disparity suggests that fine-tuning may distort learned features when data is limited. Regarding LoRA’s performance, it is observed that LoRA with a rank of 1 performs better than higher ranks, indicating that a simpler model architecture can suffice for AM tasks. Conversely, in the SM case study, finetuning from the outset yields the best results. This discrepancy can be attributed to the nuanced surface-level information in SM tasks, which requires fine-tuning for effective adaptation. Interestingly, the choice of LoRA rank varies across AM and SM case studies. A rank of 1 performs better in AM scenarios, while a rank of 8 is optimal for SM tasks. This parallels the performance trends observed between probing and finetuning. Notably, stable performance is evident from Table \ref{tab:4}, where LoRA rank 1 consistently exhibits higher performance at 10000 shots compared to probing with MLP, while also outperforming at few shots in the SM case study. The model with the poorest performance under the respective conditions is highlighted in lighter green, while the best-performing model is highlighted in darker red. Although it does not reach the performance level of finetuning with 10000 shots, LoRA with rank 1 has merely thousands of learnable parameters, comparing to the million scale of finetuning. 

\subsection{Effect of Including Task Dependent Input}
Given the widespread use of simple heuristic models for estimating manufacturability, it is natural to explore whether incorporating this information with VIRL-pretrained latent code can enhance performance further. In this section, heuristic estimations are derived for AM print time and SM cut time. After that, the impact of including such estimates as task dependent input (TDI) is examined in downstream regression. 

For estimating AM print time, a common parametric approach considers the part’s volume and surface area \cite{Hollis2001, choi2002modelling}. In this study, the math model adopted incorporates contour scanning and infill times, formulated in Eqn. \ref{eqn1}, and Eqn. \ref{eqn2}.

\begin{equation}
V_t \approx V_i + V_c + V_s + V_a \approx V_p \times IP + A_p \times w + V_s + V_a
\label{eqn1}
\end{equation}

\begin{equation}
T \approx \frac{V_t}{dV_{\text{out}}/dt} \approx (V_i + V_c) \times \frac{dV_{\text{out}}}{dt} + (V_s + V_a) \times \frac{dV_{\text{out}}}{dt} = \alpha \times (V_p \times IP + A_p \times w) + \beta
\label{eqn2}
\end{equation}

\begin{figure}[]
    \centering
    \includegraphics[width=\textwidth]{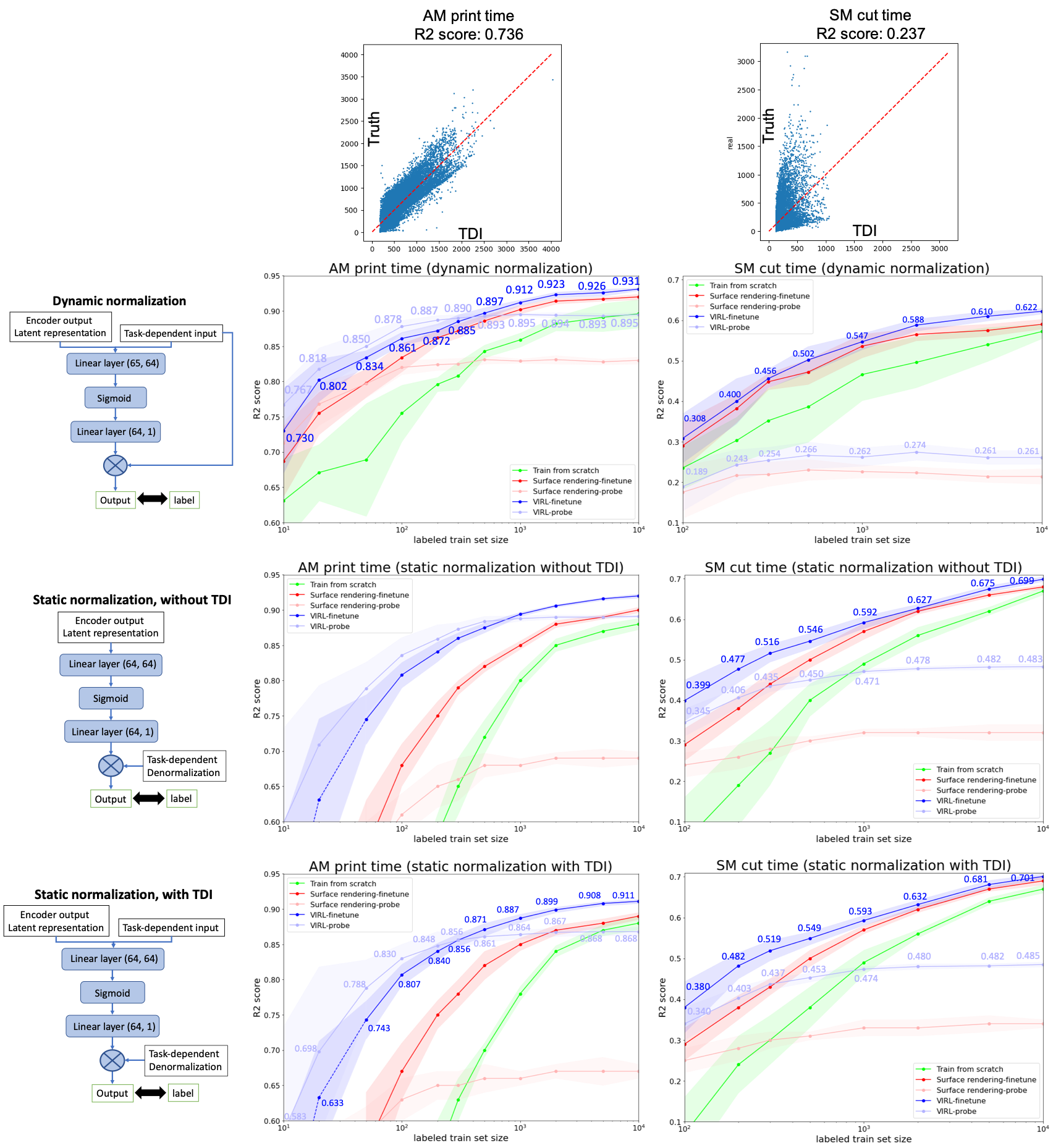} 
    \caption{Comparing model performance estimating AM/SM time with static/dynamic normalization.}
    \label{fig:11}
\end{figure}

The concept behind Eqn. \ref{eqn1} is that the total volume to be printed ($V_t$) is an approximation of the sum of infill ($V_i$), contour scanning ($V_c$), support ($V_s$), and adhesion print time ($V_a$). $V_t$ and $V_i$ can be approximated from the CAD's volume ($V_p$) and surface area ($A_p$) properties. Printing configurations, including wall thickness ($w$) and infill percentage ($IP$), are also considered.

Eqn. \ref{eqn2} represents the total print time ($T$), neglecting travel and retraction, as approximately the total volume to be printed divided by the nozzle injection rate (${dV_{\text{out}}/dt}$). By substituting known terms from Eqn. \ref{eqn1}, Eqn. \ref{eqn2} can be simplified further into a linear regression model with two unknown variables, $\alpha$ and $\beta$. The resulting R2 score and scatter plot from this parametric model, depicted in Fig. \ref{fig:11}, yield a moderate accuracy score of 0.74, as expected due to the model's simplicity.

Regarding SM cut time estimation, efforts are initially adopted to develop a heuristic approach akin to that used in AM print time, using both CAD properties and manufacturing settings. This involved dividing the subtracted volume by the estimated material removal rate derived from machine and tool bit parameters. However, the accuracy of this method is notably poor, primarily due to the challenge of discerning the varying volume removed by each tool bit. This finding aligns with previous literature indicating limited utility of parametric methods, which rely on simple regression models to estimate machining time based on subtracted volume and other geometric properties, particularly in scenarios involving multiple machining processes \cite{Armillotta2021}. Subsequently, a simple linear regression model is used with subtracted volume as the sole input. While this approach yields a more reliable estimate than previous attempts, its performance remains inferior. The resulting R2 score of 0.23, shown in Fig. \ref{fig:11}, underscores the considerable difficulty inherent in accurately estimating SM cut time compared to AM print time. 

It is found through experiment that when TDI is incorporated, dynamic normalization handles TDI effectively by dynamically multiplying the neural network’s output before comparing to the ground truth. However, when TDI is not utilized, it is preferable to normalize the data by regressing at the log domain and multiplying with a static denormalization term. The plot in Fig. \ref{fig:11} illustrates the R2 score performance on SM cut time and AM print time, with both static and dynamic normalization.

For AM print time, incorporating TDI through dynamic normalization yields significant performance improvements with limited data (\textless1000 shots), while also slightly enhancing the upper bound when there is a large dataset (10000 shots). However, dynamic normalization has a detrimental effect on SM cut time. Notably, the R2 score is substantially worse with limited data, and even with a large dataset, the performance gap difference between the two normalization methods does not diminish. In both case studies, it is important to note that including TDI in static normalization barely impacts performance. Through this experiment, it is apparent that the variance in performance could be attributed to the quality of the heuristic estimation model, as reflected in the R2 scores. The key takeaway from this experiment is twofold: if a robust heuristic estimation model is unavailable, static normalization is advisable. Conversely, if a reliable heuristic estimation model exists, dynamic normalization can significantly enhance performance, especially in scenarios with limited data. However, the use of dynamic normalization without a dependable heuristic model may lead to decreased performance. Additionally, whether TDI is included in static normalization seems to have minimal impact on overall performance. 

\subsection{Ablation Study}
\subsubsection{Effect of Modifying Hidden Layer Width and Latent Code Width}
Given the differing performance characteristics of probing with limited data and finetuning with larger datasets, Table \ref{tab:5} and Table \ref{tab:6} exclusively showcase results for probing with 100 shots using MLP, representing scenarios with limited data, and finetuning with 10,000 shots, representing scenarios with ample data. The model with the poorest performance under the respective conditions is highlighted in lighter green, while the best-performing model is highlighted in darker red.

Results indicate that increasing the latent code width above 64 yields only marginal improvements in finetuning on large datasets, while potentially resulting in deteriorated performance in probing on limited data. However, the consistent performance enhancement observed in the SM cut time case study as the latent code width increases suggests the inherently higher complexity of this particular case study, highlighting the case-specific nature of performance. On the contrary, increasing the hidden layer width does not hinder either few-shot probing or large data finetuning. Nonetheless, it is noted that performance tends to plateau when the width exceeds 1024. Thus, it is essential to recognize that the substantial increase in learnable parameters from 6.41 million to 25.4 million when opting for a width of 2048 may not yield significant advantages. 

\begin{table}[h]
\renewcommand{\arraystretch}{1.3}
\centering
\caption{VIRL-pretrained models’ performance with varying latent code width.}
\begin{tabular}{cc|c|c|c|c|c}
\multicolumn{2}{c|}{\textbf{latent code width}} & \textbf{16}   & \textbf{32}   & \textbf{64}   & \textbf{128}  & \textbf{256}  \\ \hline
\multicolumn{2}{c|}{\textbf{Encoder parameters}} &
  \textbf{6.36 M} &
  \textbf{6.38 M} &
  \textbf{6.41 M} &
  \textbf{6.48 M} &
  \textbf{6.61 M} \\ \hline
\multicolumn{1}{c|}{\multirow{4}{*}{\begin{tabular}[c]{@{}c@{}}probe\\ 100\\ shots\end{tabular}}} &
  SM time &
  {\color[HTML]{00ca5b} 0.22} &
  0.26 &
  0.35 &
  0.37 &
  {\color[HTML]{cc0000} 0.39} \\ \cline{2-7} 
\multicolumn{1}{c|}{}    & AM time     & 
{\color[HTML]{cc0000} 0.84} & 
{\color[HTML]{cc0000} 0.84} & 
0.83 & 
0.82 & 
{\color[HTML]{00ca5b} 0.81} \\ \cline{2-7} 
\multicolumn{1}{c|}{}    & AM blade    & 
0.41 & 
{\color[HTML]{cc0000} 0.47} & 
0.43 & 
0.41 & 
{\color[HTML]{00ca5b} 0.37} \\ \cline{2-7} 
\multicolumn{1}{c|}{}    & AM stress   & 
0.31 & 
{\color[HTML]{cc0000} 0.37} & 
0.31 & 
0.26 & 
{\color[HTML]{00ca5b} 0.25} \\ \hline
\multicolumn{1}{c|}{\multirow{4}{*}{\begin{tabular}[c]{@{}c@{}}finetune\\ 10000\\ shots\end{tabular}}} &
  SM time &
  {\color[HTML]{00ca5b} 0.66} &
  0.68 &
  0.70 &
  0.71 &
{\color[HTML]{cc0000} 0.72} \\ \cline{2-7} 
\multicolumn{1}{c|}{}    & AM time     & {\color[HTML]{00ca5b} 0.90} & 0.91 & 
{\color[HTML]{cc0000} 0.92} & 
{\color[HTML]{cc0000} 0.92} &
{\color[HTML]{cc0000} 0.92} \\ \cline{2-7} 
\multicolumn{1}{c|}{}    & AM blade    & {\color[HTML]{00ca5b} 0.56} & 0.60 & 0.62 & 0.62 & {\color[HTML]{cc0000} 0.65} \\ \cline{2-7} 
\multicolumn{1}{c|}{}    & AM stress   & {\color[HTML]{00ca5b} 0.52} & 0.56 & 0.58 & 0.59 & {\color[HTML]{cc0000} 0.61} \\ 
\end{tabular}

\label{tab:5}
\end{table}

\begin{table}[]
\renewcommand{\arraystretch}{1.3}
\centering
\caption{VIRL-pretrained models’ performance with varying hidden layer width.}
\begin{tabular}{cc|c|c|c|c|c|c|c}
\multicolumn{2}{c|}{\textbf{hidden layer width}} & \textbf{32} & \textbf{64} & \textbf{128} & \textbf{256} & \textbf{512} & \textbf{1024} & \textbf{2048} \\ \hline
\multicolumn{2}{c|}{\textbf{Encoder parameters}} & \textbf{9.92 K} & \textbf{32.1 K} & \textbf{113 K} & \textbf{423 K} & \textbf{1.63 M} & \textbf{6.41 M} & \textbf{25.4 M} \\ \hline
\multicolumn{1}{c|}{\multirow{4}{*}{\begin{tabular}[c]{@{}c@{}}probe\\ 100\\ shots\end{tabular}}} & SM time &
{\color[HTML]{00ca5b} 0.17} & 0.22 & 0.20 & 0.25 & 0.32 & {\color[HTML]{cc0000} 0.35} & 0.33 \\ \cline{2-9} 
\multicolumn{1}{c|}{} & AM time &
0.76 & {\color[HTML]{00ca5b} 0.69} & 0.79 & 0.81 & 0.82 & {\color[HTML]{cc0000} 0.83} & {\color[HTML]{cc0000} 0.83} \\ \cline{2-9} 
\multicolumn{1}{c|}{} & AM blade & 
0.30 & {\color[HTML]{00ca5b} 0.24} & 0.37 & 0.41 & {\color[HTML]{cc0000} 0.46} & 0.43 & {\color[HTML]{cc0000} 0.46} \\ \cline{2-9} 
\multicolumn{1}{c|}{} & AM stress & 
0.33 & {\color[HTML]{00ca5b} 0.26} & 0.34 & 0.33 & {\color[HTML]{cc0000} 0.37} & 0.31 & 0.33 \\ \hline
\multicolumn{1}{c|}{\multirow{4}{*}{\begin{tabular}[c]{@{}c@{}}finetune\\ 10000\\ shots\end{tabular}}} & SM time & {\color[HTML]{00ca5b} 0.45} & 0.47 & 0.56 & 0.63 & 0.67 & {\color[HTML]{cc0000} 0.70} & {\color[HTML]{cc0000} 0.70} \\ \cline{2-9} 
\multicolumn{1}{c|}{} & AM time & 
0.80 & {\color[HTML]{00ca5b} 0.78} & 0.85 & 0.89 & 0.90 & {\color[HTML]{cc0000} 0.92} & 0.91 \\ \cline{2-9} 
\multicolumn{1}{c|}{} & AM blade & 
0.41 & {\color[HTML]{00ca5b} 0.39} & 0.50 & 0.55 & 0.59 & {\color[HTML]{cc0000} 0.62} & 0.54 \\ \cline{2-9} 
\multicolumn{1}{c|}{} & AM stress & 
0.40 & {\color[HTML]{00ca5b} 0.39} & 0.46 & 0.51 & 0.55 & {\color[HTML]{cc0000} 0.58} & {\color[HTML]{cc0000} 0.58} \\
\end{tabular}

\label{tab:6}
\end{table}

\subsubsection{Effect of Early Stopping}
\begin{figure}[h]
    \centering
    \includegraphics[width=13cm]{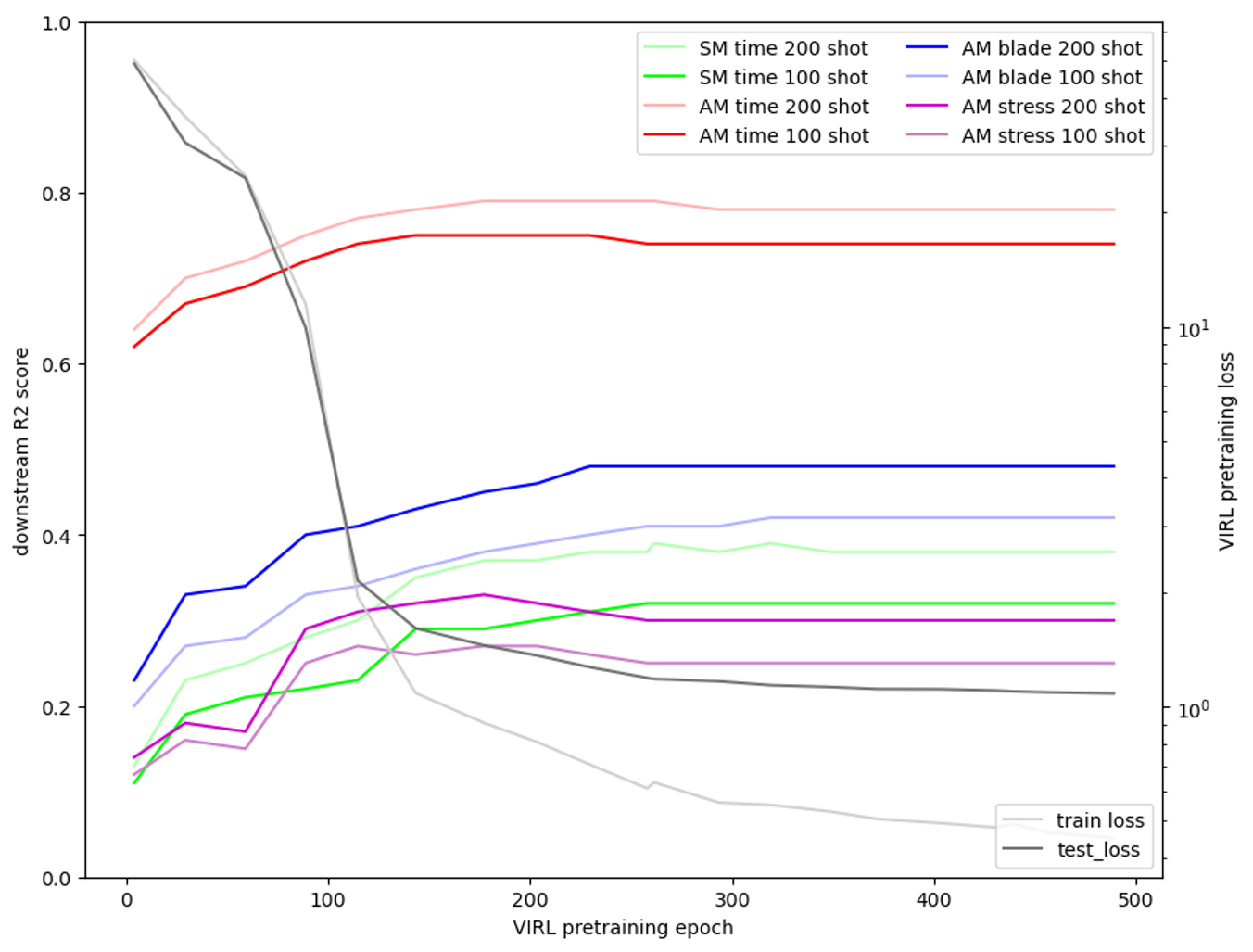} 
    \caption{Downstream performance under different pretraining stages.}
    \label{fig:12}
\end{figure}

It is also crucial to examine the embryology of VIRL \cite{Chiang2020} to determine the optimal training duration and conserve computational resources. Fig \ref{fig:12} illustrates the loss curves during VIRL pre-training alongside the performance of the four downstream tasks measured in R2 score, based on regressions from SVR on 100 and 200 shots. 

The results indicate that even when the training and testing loss have diverged, most downstream case studies still exhibit improvements in performance with further training until the point where test loss ceases to improve. Additionally, it’s noteworthy that excessive training trivially affects downstream performance. Particularly, the performance of tasks such as AM blade collision hazard and SM cut time benefits from prolonged training duration, while AM residual stress and AM printing time performance exhibit a decline with extended training. However, it is essential to highlight that this decline in performance is not substantial.

\section{Limitations and Future Work}
\label{sec:future work}
\subsection{Enhancing Dataset Realism}
First, the critical need is recognized to enrich simulation details to better represent real-world complexities. A primary goal is to introduce scalability to the geometries within the dataset, allowing one to assess whether representation learning can effectively capture properties across different scales. Second, While the orientations of CAD parts are optimized for SM cut time, future plans involve extending such pre-processing strategy to AM simulations. This is because specific rotations may optimize factors such as stress, blade interference, and print time. In addition, although the SM simulation presented in this research is generalized to output a sequential process that can successfully machine most parts in Fusion360 segmentation repository, it has neglected or idealized several processes including jig setup time, avoiding collision, and exploring alternative cutting strategies beyond surface geometry adherence, such as sweeping or depth-first cutting. 

\subsection{Enhancement of Manufacturing Parameter Diversity}
In AM, diversification entails exploring variations in laser welding temperatures, employing different types of printing machines, and experimenting with various materials. Meanwhile, in the context of SM, diversification encompasses utilizing a wide array of dimensions and shapes for tool bits, cutting materials of different compositions, and incorporating alternative machining techniques such as turning and sheet metal processing. By expanding the scope of manufacturing parameters, future work aims to enhance the robustness and applicability of pre-trained models, enabling them to accommodate a broader range of scenarios. 

\subsection{Model Exploration for Enhanced Generalization}
While direct inference from Boundary Representation (BRep) offers speed advantages by circumventing voxelization, its performance in simulating physics fields falls short of expectations, according to Fig. \ref{fig:8}. In contrast, leveraging simulation results derived from voxelized mesh Finite Element Analysis (FEA), the potential of whether Voxel Convolutional Neural Networks (CNNs) can deliver superior outcomes is investigated. Findings reveal that FeatureNet \cite{zhang2018featurenet}, a voxel-based CNN model, does demonstrate notable improvements in stress prediction and recoater interference percentage estimation over VIRL-pretrained GCN, with the same scale of learnable parameters. The few-shot performance comparsion is shown in Table \ref{tab:7}. However, its superiority is less apparent in predicting AM print time and absent in forecasting SM cut time. This highlights the significance of task specificity, and indicates that despite longer deployment times due to voxelization, CNNs, with their robust inductive bias, may outperform BRep-based GCN models even after pre-training. Nevertheless, the fact that VIRL pre-training significantly improved BRep-based GCN models has demonstrated VIRL as a viable representation learning strategy, capable of potentially adapting to various encoder types and input modalities. The prospect of exploring such versatility holds promise for future research endeavors. Forthcoming efforts will be dedicated to constructing a more versatile framework by exploring diverse combinations of input modalities and encoder-decoder structures. 

\begin{table}[H]
\centering
\caption{Few-shot performance comparison of VIRL and FeatureNet.}
\renewcommand{\arraystretch}{1.4}
\begin{tabular}{p{1.5cm}|p{2cm}|p{2cm}|p{2cm}|p{2cm}|p{2cm}|p{2cm}|p{2cm}|p{2cm}}
\multicolumn{1}{c|}{\textbf{Task}} &
  \multicolumn{2}{c|}{\textbf{SM cut time}} &
  \multicolumn{2}{c|}{\textbf{AM print time}} &
  \multicolumn{2}{c|}{\textbf{AM blade collision}} &
  \multicolumn{2}{c}{\textbf{AM residual stress}} \\ \hline
\multicolumn{1}{c|}{\textbf{Shots}} &
\multicolumn{1}{c|}{\textbf{\hspace*{3.0mm} 100\hspace*{3.5mm} }}  & 
\multicolumn{1}{c|}{\textbf{\hspace*{3.0mm} 200\hspace*{3.5mm} }} &
\multicolumn{1}{c|}{\textbf{\hspace*{3.0mm} 100\hspace*{3.5mm} }}  & 
\multicolumn{1}{c|}{\textbf{\hspace*{3.0mm} 200\hspace*{3.5mm} }} &
\multicolumn{1}{c|}{\textbf{\hspace*{3.0mm} 100\hspace*{3.5mm} }}  & 
\multicolumn{1}{c|}{\textbf{\hspace*{3.0mm} 200\hspace*{3.5mm} }} &
\multicolumn{1}{c|}{\textbf{\hspace*{3.0mm} 100\hspace*{3.5mm} }}  & 
\multicolumn{1}{c}{\textbf{200}} \\ \hline
\multicolumn{1}{c|}{Ours}  & 
\multicolumn{1}{c|}{0.40} & 
\multicolumn{1}{c|}{0.48} & 
\multicolumn{1}{c|}{0.84} & 
\multicolumn{1}{c|}{0.86} & 
\multicolumn{1}{c|}{0.38} & 
\multicolumn{1}{c|}{0.50} & 
\multicolumn{1}{c|}{0.30} & 
\multicolumn{1}{c}{0.37} \\ \hline
\multicolumn{1}{c|}{FeatureNet} &
  \multicolumn{1}{c|}{\textless{}0.1} &
  \multicolumn{1}{c|}{\textless{}0.1} &
  \multicolumn{1}{c|}{0.83} &
  \multicolumn{1}{c|}{0.85} &
  \multicolumn{1}{c|}{0.48} &
  \multicolumn{1}{c|}{0.59} &
  \multicolumn{1}{c|}{0.37} &
  \multicolumn{1}{c}{0.45} \\ 
\end{tabular}

\label{tab:7}
\end{table}

\section{Conclusion}
\label{sec:conclusion}
To swiftly estimate manufacturability properties in early-stage design, this research attempts to alleviate the limitations imposed by time-consuming CAM simulations and the challenges they present to supervised learning methods due to the lack of generated data. Our novel approach, Volume-Informed Representation Learning (VIRL), involves the development of a self-supervised representation learning task applied to unlabeled CAD geometries to provide an encoder with spatial understanding. Through the four manufacturability case studies created through CAM simulations, it is shown that VIRL pretraining not only enhances few-shot capabilities—outperforming both training from scratch and previously proposed surface rendering pretraining—but also elevates the upper bound given substantial data. 

Additionally, this research uncovers that finetuning VIRL-pretrained model has a negative impact on all three AM tasks when data is limited, but proves advantageous for SM machining time estimation. Moreover, the effectiveness of Low-rank adaptation (LoRA), a method that strikes a balance between probing and finetuning, is investigated. LoRA demonstrates consistent performance similar to probing with limited data, while achieving a higher upper bound than probing as the dataset size increases, all without the computational overhead of finetuning. Furthermore, the effect of including heuristic task dependent input in downstream tasks is studied by comparing SM cut time and AM print time estimation. Results indicate that when a robust heuristic estimation is unavailable, deploying a pretrained model by static normalization delivers consistent performance when there are more than 1000 data, and is therefore advisable. Conversely, if a reliable heuristic estimation model exists, dynamic normalization by multiplying the model’s output with such task dependent input can significantly enhance performance, especially in scenarios with limited data (100 samples). 

From an application standpoint, VIRL has the potential of liberating designers from the constraints of simulation execution, while from a research perspective, it introduces significant flexibility to data-driven methodologies by providing a valuable initialization for supervised learning methods. Moreover, it has the potential to relieve data curators from the burdens associated with collecting extensive datasets.

\section*{Acknowledgments}
This work was supported in part by the National Science Foundation under grant Award CMMI-2113301 and the Pennsylvania Infrastructure Technology Alliance.

\section*{Data Availability}
Our dataset with four manufacturability labels are available on Github:
\href{https://github.com/parksandrecfan/VIRL}{https://github.com/parksandrecfan/VIRL}

\bibliographystyle{unsrt}  
\bibliography{references}  

\end{document}